# Category Theory as a Foundation for Soft Robotics

**Hayato Saigo[1*], Makoto Naruse[2*], Kazuya Okamura[3], Hirokazu Hori[4] and Izumi Ojima[5]**

1 Nagahama Institute of Bio-Science and Technology, 1266 Tamura-Cho, Nagahama, Shiga 526-0829, Japan

2 Network System Research Institute, National Institute of Information and Communications Technology, 4-2-1 Nukui-kita, Koganei, Tokyo 184-8795, Japan

3 Graduate School of Informatics, Nagoya University, Furo-cho, Chikusa-ku, Nagoya, Aichi 464-8601, Japan

4 Interdisciplinary Graduate School, University of Yamanashi, Takeda, Kofu, Yamanashi 400-8511, Japan

5 Shimosakamoto, Otsu, Shiga 520-0105, Japan

* Correspondence: Hayato Saigo (email: harmoniahayato@gmail.com) and Makoto Naruse (email: naruse@nict.go.jp)

**Abstract**

Soft robotics is an emerging field of research where the robot body is composed of compliant and soft materials. It allows the body to bend, twist, and deform to move or to adapt its shape to the environment for grasping, all of which are difficult for traditional hard robots with rigid bodies. However, the theoretical basis and design principles for soft robotics are not well-founded despite their recognized importance. For example, the control of soft robots is outsourced to morphological attributes and natural processes; thus, the coupled relations between a robot and its environment are particularly crucial. In this paper, we propose a mathematical foundation for soft robotics based on category theory, which is a branch of abstract math where any notions can be described by objects and arrows. It allows for a rigorous description of the inherent characteristics of soft robots and their relation to the environment as well as the differences compared to conventional hard robots. We present a notion called the category of mobility that well describes the subject matter. The theory was applied to a model system and analysis to highlight the adaptation behavior observed in universal grippers, which are a typical example of soft robotics. This paper paves the way to developing a theoretical background and design principles for soft robotics.

**Keywords: soft robot, category theory, natural transformation, system modeling, autonomous adaptation**

# 1. INTRODUCTION

Soft robotics is being intensively studied to overcome the difficulties of traditional robots made from a rigid underlying structure and to create new value by exploiting the intrinsically soft and/or extensible material (Rus et al., 2015; Trivedi et al., 2008; Pfeifer et al., 2007). Conventional hard robots with rigid bodies require precision control of a number of discrete actuators for links and joints; they perform well at repeating well-defined motions for massive manufacturing but have severe difficulties in dealing with an uncertain environment. In contrast, soft robots employ compliant materials that provide abilities such as bending, twisting, or deforming of their bodies to adapt their shape to the environment (Rus et al., 2015). Experimental demonstrations including the multi-gait soft robot (Shepherd et al., 2011), universal gripper (Brown et al., 2010), and octopus robot (Laschi et al., 2012) have highlighted the deformability of their bodies. Experimental implementations of more abilities such as logic gates in soft materials (Wehner et al., 2016), camouflage, and displays (Morin et al., 2012) have also been studied for soft robots.

Soft robotics is inspired by biology, such as the muscle–tendon complex, skin sensors, and retina observed in living organisms (Rus et al., 2015; McEvoy et al., 2015; Pfeifer et al., 2012). Pfeifer *et al.* emphasized the aspect of *embodiment*: behavior is not only controlled by the brain (or signal processing units) but is the result of the reciprocal dynamical coupling of the brain, body, and environment (Pfeifer et al., 2012). Part of the control of soft robots is outsourced to morphological and material properties; hence, there is a demand for novel design principles that can be applied to soft robotics (Pfeifer et al., 2012).

However, the theoretical background of soft robotics is limited to a few studies in the literature, which are introduced below. This is in contrast to hard robots, where a vast and deep theoretical foundation is provided, ranging from conventional and modern control theories (de Wit et al., 2012) to architectural insights such as subsumption architecture (Brookes, 1986). Regarding the theory for soft robots, Pfeifer *et al.* presented an architectural analysis of the controller, mechanical system, sensory system, and task environment (Pfeifer et al., 2007); Brown *et al.* performed ad hoc but detailed dynamical system modeling of grippers (Brown et al., 2010); Hauser *et al.* examined the morphological computation of compliant bodies (Hauser et al., 2011); and Ay *et al.* performed information theoretic analysis of sensorimotor loop by utilizing transfer entropy to examine the causal structures therein (Ay et al., 2014). All of these early theoretical studies shed light on the unique aspects and properties of soft robotics. In this study, we contribute to the literature of soft robotics via category theory.

The challenges for developing a theoretical basis for soft robotics stem from a variety of unique attributes that are difficult to describe with conventional frameworks. First, soft robots have a huge number of degrees of freedom. For example, the number of elements in a deformable body can be on the scale of Avogadro's number if the precision needs to be at a very fine scale. Second, soft robots need to adapt to the environment and its uncertainty. Environmental information, such as the shape of the object of interest to be grasped, is difficult to acquire precisely prior to physical interactions unless high-precision measurement systems are assumed. Accommodating a diverse range of environments remains to be addressed.

In this paper, we propose an approach to soft robotics based on category theory, which is a branch of mathematics that simplifies all mathematical notions into objects and arrows (Mac Lane, 1971; Awodey, 2010; Simmons, 2011). Recently, categorical understanding of decision making and solution searching has been demonstrated (Naruse et al., 2018, 2017), where the uncertain environmental entities are described and the underlying mechanisms are characterized by utilizing the theorems and axioms known in the triangulated category. In this study, we adapted the notions of functors and natural transformation in category theory (Mac Lane, 1971; Awodey, 2010; Simmons, 2011) to soft robotics by introducing the notion of the category of mobility. The unique attributes of a soft robot are matched to the framework of the category of mobility in that it encompasses the relation between two systems (robot and target) and its evolutions.

To facilitate intuitive understanding, we discuss a grasping problem as an example application of the theory. This is schematically shown in Figure 1. The difficulty of conventional hard robotics lies in the precision control of rigid joints to grasp the object of interest, whereas the soft robotics approach with the universal gripper outsources the adaptation to the object to numerous small particles (in this case, coffee beans) contained in the glove at the end of the arm (Brown et al., 2010). With regard to grasping theory, Brown *et al.* presented a mechanical analysis for the universal gripper (Brown et al., 2010), while Higashimori *et al.* discussed a pre-shaping strategy for hard robot hands (Higashimori et al., 2005). In contrast, we focus on the universality and autonomous adaptation of soft robots. A qualitative agreement is observed between the theoretical prediction and the numerical experiment. It should also be emphasized that the notion of scale-dependent fitting measure between the object and the gripper, to be introduced precisely in the experiment section, is inspired by categorical thinking. With such figures, the explosion of the number of possible states and the inter-state transitions are clearly observed when overly small scales are considered: such an aspect is accommodated in the theory by means of natural transformation.

The remainder of this paper is organized as follows. Section 2 presents an approach based on category theory to soft robots. Based on the theory, Section 3 presents the modeling and analysis of a universal gripper that serves to highlight the autonomous adaptation process with a massive number of elements. Section 4 concludes the paper.

## 2. RESULTS: THEORY

### 2.1. Category Theory as a Foundation of Soft Robotics

We introduce fundamental concepts in category theory such as *category*, *functor*, and *natural transformation*. To formulate the notion of a category of mobility, we first need the notion of a category. To define the notions of soft and effectively soft robots, we need the notions of categorical isomorphism and categorical equivalence. These are defined in terms of the functor and natural transformation. For a more detailed introduction to category theory, see Mac Lane (1971) for example.

### 2.1.1. Categories

A category is a network formed from composable arrows that intertwine with objects. The objects can be considered to represent some phenomena, while the arrows show a transformation or process between these phenomena. A category is a system consisting of objects and arrows that satisfies the following four conditions:

1. Each arrow $f$ is associated with two objects $\mathrm{dom}(f)$ and $\mathrm{cod}(f)$, which are called the domain and codomain, respectively. When $\mathrm{dom}(f) = X$ and $\mathrm{cod}(f) = Y$, the following can be expressed:
$$f : X \to Y \tag{1}$$
or
$$X \xrightarrow{f} Y. \tag{2}$$
The direction of the arrow does not need to be limited to from left to right; if convenient, it can be from bottom to top, from right to left, etc. A subsystem of the category built up with these arrows and objects is called a diagram.

2. Assume that there are two arrows $f$ and $g$ such that $\mathrm{cod}(f) = \mathrm{dom}(g)$:
$$Z \xleftarrow{g} Y \xleftarrow{f} X. \tag{3}$$
Then, there is a unique arrow called the composition of $f$ and $g$:
$$Z \xleftarrow{g \circ f} X. \tag{4}$$

3. Assume the associative law for the following diagram:

$$W \xleftarrow{h \circ g} Y$$
$$h \nwarrow \quad g \swarrow \quad \nwarrow f$$
$$Z \xleftarrow{g \circ f} X \quad . \tag{5}$$

Then, we can assume the following:

$$(h \circ g) \circ f = h \circ (g \circ f) . \tag{6}$$

When any compositions of arrows with the same codomain and domain are equal, the diagram is called commutative.

4. The last condition is the unit law. For any object $X$, there exists a morphism $1_X : X \to X$ called the identity of $X$ such that the following diagram is commutative for any $f: X \to Y$:

$$\begin{array}{c} X \\ f \swarrow \quad \nwarrow 1_X \\ Y \xleftarrow{f} X \\ 1_Y \searrow \quad \swarrow f \\ Y \end{array} \quad . \tag{7}$$

In other words, $f \circ 1_X = f = 1_Y \circ f$.

Because of the natural correspondence between objects and their identities, we may identify the objects as identities. In other words, we may consider the objects as just the special morphisms. In the rest of this paper, we may adopt this viewpoint without notice.

**Definition 1 (Category)** *A category is a system composed of two kinds of entities called objects and arrows that are interrelated through the notions of the domain and codomain and is equipped with a composition and identity to satisfy associative law and unit law.*

There is so many examples of categories in mathematics; almost everything in mathematics can be formulated in terms of categories. For example, the category **Set** contains sets as objects and mappings as arrows. Another example is the category of propositions, where objects are propositions and arrows are proofs. Any partially ordered set can be considered as a category; objects are elements of the set, and arrows are the order relation (e.g., larger than) between elements. By definition, this category has at most one arrow between two objects. There is a unique arrow if the codomain is larger than the domain and no arrow otherwise. If we consider a category where the objects are propositions and arrows indicate provability, then the category has at most one arrow between two objects. In general, by considering arrows to indicate "-ability" or "-itivity," we can obtain the same kind of categories. We define another universal example in the next subsection: the category of mobility.

## 2.2. Isomorphisms

Once a category given, we can define the sameness between different objects in terms of isomorphism:

**Definition 2 (Isomorphism)** *An arrow f: $X \to Y$ is called an isomorphism if there is an arrow g: $Y \to X$ that makes the following diagram:*

$$
\begin{array}{c}
Y \\
{}^{1_Y}\swarrow \quad \searrow^{g} \\
Y \xleftarrow{f} X \\
{}_{g}\searrow \quad \swarrow_{1_X} \\
X
\end{array}
\tag{8}
$$

*In other words,*

$$g \circ f = 1_X, f \circ g = 1_Y. \tag{9}$$

*When there is an isomorphism between X and Y, they are called isomorphic and denoted by $X \cong Y$.*

Two different objects are considered essentially the same in the category if they are isomorphic. Because they are connected by an isomorphism, if one is in a certain diagram, the other is in a completely similar diagram. This is why we can count things with both fingers and stones; a set of five fingers and set of five stones are isomorphic in the category of sets and functions. The notion of "five" is obtained by recognizing such isomorphisms.

In summary, this means that, every time one category is postulated, a suitable sameness called isomorphism is determined for objects that are contacted by reversible arrows in the category. Here, we present the basic structure for elucidating the sameness between obviously different phenomena.

## 2.3. Category of Mobility

In this world, it is impossible for any phenomenon to be completely identical with any other phenomenon. Therefore, there is no repeatable phenomenon in an absolute sense. This is the undeniable nature of things. It is only possible for us to discuss a law in the form of saying "the same thing can be said about similar phenomena" when we set up an equivalent relation between different phenomena.

First, we regard each phenomenon that is restlessly changing as approximately constant when in the same system with respect to some equivalence relation. Furthermore, we consider the pattern of the relationship between the system and environment, which is fundamentally unique and irrepealable, under some criteria of equivalence.

The concepts of the system and interface between the system and environment, under certain equivalence relations, should be essential ingredients in every science. In physics, these concepts are formulated in terms of algebras of quantities (i.e., observables), and states are the expectation functionals of the algebras. The pair of the algebra of quantities and its state provide a general framework to describe the statistical law for a certain system in a certain relationship with the environment. The state connects the quantities themselves and their values in a spectrum and reflects the condition based on the relationship between the system and environment. The state connects the past and future; it can be defined through the equivalence between preparation processes and provides a basis for discussing the possible transition of the relationship between the system and environment.

The concept of the category of mobility is based on the idea that the state at a moment is an effect caused by hidden dynamics while at the same time causing the future development of the dynamics. In general, such a development cannot be expected to be deterministic. Then, we can consider all possible transitions between states, starting from the given initial state $\varphi$, to become a category that contains all possible future states as objects and transitions as arrows. We call this the category of the state transition from $\varphi$. The category contains full information of future possibilities. However, for the purpose of the present paper and related works, it is sufficient to consider a rough sketch: the category of mobility from $\varphi$, which is denoted as **Mob**$_\varphi$. This category contains all future states that can be realized in the not-so-long term, and its arrows indicate transitivity between states in the not-so-long term. Here, the seemingly ambiguous term "not-so-long term" has the same implication as for the formulation of the zeroth law of thermodynamics. It simply stresses the importance of the time scale and dependence on the basic hypothesis, such as *ceteris paribus* (i.e., the state of the environment can be seen as constant).

**Definition 3 (Category of Mobility)** *Let $\varphi$ be the initial state of a system and suppose that the state $\phi$ of the environment around that system can be considered as constant. The category of mobility **Mob**$_{\varphi,\phi}$ of the system with the initial state $\varphi$ has future states as objects and possible state transitions in state $\phi$ of the environment as arrows. We denote **Mob**$_{\varphi,\phi}$ as **Mob**$_\varphi$ when there is no worry of confusion.*

The notion of the category of mobility provides a theoretical framework for soft robots. It can be used to define hard, soft, and effectively soft robots together with the basic notions of category theory, which are introduced in the next subsection.

## 2.4. Functor and Natural Transformation

A functor is defined as a structure-preserving correspondence of two categories:

**Definition 4 (Functor)** *A correspondence F from C to D that maps each object/arrow in C to a corresponding object/arrow in D is called a functor if it satisfies the following conditions:*

1. *It maps $f: X \to Y$ in C to $F(f): F(X) \to F(Y)$ in D.*
2. *$F(f \circ g) = F(f) \circ F(g)$ for any (composable) pair of f and g in C.*
3. *For each X in C, $F(1_X) = 1_{F(X)}$.*

In short, a functor is a correspondence that preserves diagrams or, equivalently, the categorical structure. A functor is a very universal concept. All processes expressed by words such as recognition, representation, construction, modeling, and theorization can be considered to be the creation of functors. Once the notion of a functor is established, it is natural to introduce the concepts of the category of categories and isomorphism of categories.

**Definition 5 (Category of categories)** *The category of categories, which is denoted by **Cat**, is a category whose objects are categories and arrows are functors. The isomorphism of categories is defined as the isomorphism in **Cat**, i.e., invertible functors. Two categories are categorically isomorphic if they are isomorphic as objects in **Cat**. In other words, two categories C and D are categorically isomorphic if there is a pair of functors F, G such that $G \circ F = Id_C$ and $F \circ G = Id_D$, where $Id_C$ and $Id_D$ are the identity functors of C and D.*

The notion of the isomorphism of categories seems to be quite natural, but the concept of "essentially the same" categories is known to be too narrow for formulations in mathematics. To define such essential sameness in mathematical terms, which is called the equivalence of categories, we need the central notion of category theory: natural transformation.

**Definition 6 (Natural transformation)** *Let F and G be functors from category C to category D. The correspondence t is called a natural transformation from F to G if it satisfies the following conditions:*

1. *t maps each object X in C to the corresponding arrow $t_X$: F(X) → G(X) in D.*

2. *For any f: X → Y in C,*

$$t_Y \circ F(f) = G(f) \circ t_X. \tag{10}$$

For the natural transformation, we use the notation $t: F \Rightarrow G$. The second condition above is depicted as follows:

$$
\begin{array}{c|cc}
 & Y & \xleftarrow{f} & X \\
\hline
F & F(Y) & \xleftarrow{F(f)} & F(X) \\
\Downarrow & t_Y \downarrow & & \downarrow t_X \\
G & G(Y) & \xleftarrow[G(f)]{} & G(X)
\end{array}
\tag{11}
$$

The upper-right part denotes the arrow in C, and the bottom-left part denotes the arrow in D. The second condition in the definition for natural transformation means that the above diagram is commutative.

It is easy to see that the functors from C to D and natural transformation between them become a category: the functor category from C to D.

**Definition 7 (Functor category)** *The functor category from C to D, which is denoted as* Fun(C, D)*, is the category whose objects are functors from C to D and arrows are natural transformations between them. Isomorphism in the functor categories, i.e., invertible natural transformations, is called natural equivalence. Functors from C to D are called naturally equivalent if they are isomorphic as objects in* Fun(C, D).

Because any category can be considered as a subsystem of some functor category in some sense, all kinds of sameness that we can define are actually formulated in terms of natural transformations, especially natural equivalence.

For natural equivalence, which is the isomorphism between functors, we introduce the notion of the equivalence of categories. This is the functor that represents the essential sameness between categories.

**Definition 8 (Equivalence of categories)** *A functor F from C to D is called an equivalence of categories if there is a functor G from D to C such that $G \circ F$ is naturally equivalent to $Id_C$ and $Id_D$. Two categories C and D are categorically equivalent if there is an equivalence of categories. In other words, two categories C and D are categorically equivalent if there is a pair of functors F and G such that $G \circ F \cong Id_C$ and $F \circ G \cong Id_D$.*

One remark here is that isomorphism and categorical equivalence are categorical counterparts of homeomorphism and homotopy equivalence.

## 2.5. Control

To define and analyze the notion of soft robots, we begin with the fundamental notion of control of composite systems by its subsystems.

The relationships between a composite system and its subsystems are seemingly simple. Once we consider the control phenomena between subsystems and component systems, a fundamental relationship of duality become clear. On the one hand, the composite system trivially includes subsystems. On the other hand, the subsystem determines, at least statistically, the full systems. This dynamic duality of acting/acted is well-modeled from the viewpoint of category theory. This provides a framework to treat the categories of morphisms of any system as entities on equal footing: small or large, subsystem or supersystem.

The relationship between systems is modeled by functors between categories of mobility. As an important example, consider the subsystem $S_0$ of the system $S$, where $M_{\varphi_0}$ and $M_{\varphi}$ are the corresponding categories of mobility. Then, there should be projection functors $P_0 : M_{\varphi} \to M_{\varphi_0}$ that send the states of $S$ to the corresponding states of $S_0$, especially $\varphi$ to $\varphi_0$. Then, we can define the notion of control as follows:

**Definition 9 (Control)** *Let $S_0$ and $S_1$ be systems, $S$ be the composite system, and $M_{\varphi_0}$, $M_{\varphi_1}$, $M_{\varphi}$ denote their respective categories of mobility. We denote $P_i : M_{\varphi} \to M_{\varphi_i}$ sending $\varphi$ to $\varphi_i$ (i = 0, 1) as the projection functors and $s_0 : M_{\varphi_0} \to M_{\varphi}$, which is a section of $P_0$, as a functor that satisfies $P_0 \circ s_0 = \mathrm{Id}_{M_{\varphi_0}}$. The system $S_0$ controls $S_1$ through $s_0$ when the state of the system $S_0$ is $\eta$. Then, the state of the system $S_1$ is $P_1 \circ s_0(\eta)$. If there is such $s_0$, then $S_0$ controls $S_1$.*

One remark on the notion of a composite system is that it includes both the body of the robot and the body of the object under study.

## 2.6. Hard and Soft Robots

Robots control target systems by making themselves subsystems of the composite system with the target systems. The notion of hardness/softness, which is central to soft robotics, can be defined as follows.

**Definition 10 (Hard robot)** *A robot is hard when the category of mobility of the composite system of the robot and target entity is isomorphic with the category of mobility of the robot during interaction.*

We emphasize the term "isomorphic," i.e., there is the invertible functor between them. This means a coherent one-to-one correspondence between the state and transitivity of states of the composite system and robots during interaction, as schematically shown in Figure 2A. In other worlds, the robot controls the state of the composite system deterministically. In contrast, we define the notion of soft robots by replacing "isomorphic" with "categorically equivalent":

**Definition 11 (Soft robot)** *A robot is soft when the category of mobility of the composite system of the robot and target entity is categorically equivalent to the category of mobility of the robot during interaction.*

We here emphasize that the difference between a hard robot and a soft robot has been successfully defined via exact mathematical notions as the difference in the sameness given by isomorphic and categorical equivalence. Consequently, the categorical equivalence can provide indeterminacy in terms of the control. In other words, as long as the categorical equivalence is satisfied, potentially multiple, in some cases huge, physical states yield the same meaning. In certain contexts, this provides much power for the finding the best or approximately the best way to control, as schematically illustrated in Figure 2B. In contrast to hard robots, which eliminate the degrees of freedom of composite systems, soft robots make use of such degrees of freedom as some intelligence of nature. In short, soft robotics can be considered as a powerful generalization of conventional robotics based on natural intelligence.

However, not all soft robots are effective because overly soft robots will not work for detailed control. In that sense, there is a tradeoff between finding and keeping the (approximately) best way of control. In the next subsection, we define the notion of effectively soft robots.

## 2.7. Effectively Soft Robots

What kind of soft robots are effective? To reflect the tradeoff between finding and keeping, we define the notion of *effectively soft robots* as those that are soft at first and hard at the end. More precisely, we define the effectiveness of soft robots based on the notion of critical states for this soft–hard transition:

**Definition 12 (Critical state)** *A state $\varphi_C$ in the category of mobility of the composite system of a robot and target entity is called a critical state for the soft-hard transition if the category of mobility from $\varphi_C$, $M_{\varphi_C}$, is isomorphic to the category of mobility of the target entity from $\varphi_C$, $M_{P(\varphi_C)}$, where P denotes the projection functor.*

**Definition 13 (Effective)** *A category of mobility is effective if there is an arrow from any state to some critical state for the soft–hard transition.*

**Definition 14 (Effectively soft robot)** *A soft robot is effectively soft if the category of mobility contains critical states.*

The notion of effectively soft robots provides a new idea for the powerfulness of soft robots.

## 2.8. Universal gripper as an effectively soft robot

We focused on the universal gripper (Brown et al., 2010) as a typical example of an effectively soft robot. The simplest version of the universal gripper consists of a vacuum machine and small bag containing coffee beans (Brown et al., 2010). The softness is provided by the mobility of the coffee beans and flexible shape of the small bag. The softness allows the gripper to find the best shape for grasping. However, not all soft robots are effective. What makes the universal gripper an effectively soft robot?

One answer is the vacuum machine because the bag cannot keep its best shape without its help. However, there is another factor: the size of coffee beans.

It is natural to imagine that a smaller size means that the beans have higher mobility. In other words, the category of mobility becomes rich with arrows, and the robot becomes softer. However, the critical states become scarce, so the category of mobility becomes less effective. If this reasoning is correct, there should be optimal size of coffee beans that make the universal gripper effectively soft. In the next section, we present numerical simulations performed to investigate this aspect.

In addition, the notion of categorical equivalence in the definition of a soft robot clearly conveys that fluctuations or deformations are inherently accommodated. Furthermore, in the case of the universal gripper, concerning the fact that the composite system is made by the gripper and the target, the adaptation processes will differ depending on the shape of the target to be handled. In the meantime, we will show that the number of states and transitions is indeed huge if we consider the smaller physical scale involved.

## 3. DISCUSSION: MODELING AND ANALYSIS

To demonstrate the autonomous adaptation of soft robots manifested by the category of mobility as well as quantitatively present the prediction and indication by the theory of effectively soft robot, we present a model system and analysis of a universal gripper (Brown, et al. 2010) as a typical instance of a soft robot. As introduced in the previous section, the gripper contains a number of elemental particles. The purpose of the numerical considerations herein is to demonstrate autonomous adaptation of the gripper to the shape of the object by rearrangements of the internal particles, which correspond to the richness of the transitions of states. In addition, the notion of effectively soft robot implies that an overabundance of transitions will result in faults of intended functionality: we will examine the impacts of physical scale in the adaptation processes.

The object to be grasped is depicted by a one-dimensional surface profile, as schematically shown in the lower side in Figure 3A, while a universal gripper is represented by an array of particles arranged in an orderly manner in the upper side of Figure 3A. Once the gripper approaches, touches, and harnesses the object, the particles in the glove of the gripper are rearranged, as schematically shown in Figure 3B. During the transformation from the initial state to the final state, the particles in the glove move from one place to another, as illustrated by the arrows in Figure 3B.

Note that each particle in the gripper is not controlled individually; the particles can freely move but are subjected to the constraint that the total volume, or total number of particles in the glove, is constant.

To highlight such a mechanism, we present the following hierarchical model. Suppose that the surface profiles of the object (i.e., TARGET) and gripper (i.e., GRIPPER) are given by *TARGET*($x$) and *GRIPPER*($x$), respectively. The relative difference between GRIPPER and TARGET is denoted by $h(x)$. We equate grasping to an autonomous modification of the shape of GRIPPER to TARGET because the primary interest is the statistical behavior of particles rather than the mechanical dynamics for lifting objects, such as that studied in Brown et al., 2010.

The relative difference $h(x)$, which is schematically depicted in Figure 3C-i, can be observed at a coarser scale (denoted by Scale *C*), as shown in Figure 3C-ii. In contrast, the detailed differences at a finer scale (Scale *F*) are presented in Figure 3C-iii.

The flow of particles in the glove of the gripper may occur in a region where the difference between the target and gripper is more evident. To quantify such a property, we introduce the following scale- and position-dependent fitting measure:

$$R_P^{(S)} = \left| h_P^{(S)} - \frac{h_L^{(S)} + h_R^{(S)}}{2} \right| \tag{12}$$

where $h_P^{(S)}$ denotes the height of the difference between GRIPPER and TARGET averaged over the unit at the scale $S$ (defined below) at the position $P$, while $h_L^{(S)}$ and $h_R^{(S)}$ represent the average heights of the left- and right-hand-side neighbors, respectively, of $P$. Suppose that TARGET (and GRIPPER) spans a horizontal length of $X$ (Figure 3A). Further assume that the finest scale of the horizontal resolution is given by $X/2^N$, namely, the number of pixels is given by $2^N$. For simplicity, we can assume that $X$ is also given by the power of 2. When the size of a single area is given by $L = 2^S$, there are $X/2^S$ areas in total at the corresponding scale $S$. More specifically, the average height in an area specified at the scale $S$ and position $x$ is given by

$$h^{(S)}(x) = \sum_{m=1,\cdots,L} h^{(0)}(2^S \times (x-1) + m)/2^S \tag{13}$$

where $h^{(0)}(1), \ldots, h^{(0)}(2^N)$ are the height information at the finest scale ($S = 0$).

The movement of particles between adjacent areas is autonomously induced at locations where the scale- and position-dependent metric $R_P^{(S)}$ gives the maximum value. Accordingly, we implement the following model system dynamics:

[STEP 1] Calculate $R_P^{(S)}$ (equation (12)) with respect to TARGET and the present shape of GRIPPER. Here, the scale $S$ ranges from $S_{min}$ to $S_{max}$.

[STEP 2] Find the scale and position that maximize $R_P^{(S)}$.

[STEP 3] Decrease the height of the corresponding area of GRIPPER by a unit if the sign of the content of equation (12), $h_P^{(S)} - (h_L^{(S)} + h_R^{(S)})/2$, is positive. This corresponds to a situation where the particles contained in the corresponding area, of which the number is $2^S$, are flowing out to the neighboring areas. Similarly, increase the height of the corresponding area of the

GRIPPER by a unit if the sign of the content of equation (12), $h_P^{(S)} - (h_L^{(S)} + h_R^{(S)})/2$, is negative. This corresponds to a situation where the particles contained in the corresponding area are flowing into the neighboring areas.

[STEP 4] Because of the flow of particles getting out of or getting into neighboring areas in step 3, the heights of the neighbors $h_L^{(S)}$ and $h_R^{(S)}$ increase or decrease. Here, we assume that half of the particles ($2^{S-1}$) go to the left side, and the other half ($2^{S-1}$) go to the right side. There are $2^S$ locations for the particles to be settled in the left or right areas. We randomly choose $2^{S-1}$ positions within such areas of the left- and right-hand-side neighbors, respectively, and reconfigure the height accordingly.

[STEP 5] Repeat steps 1–4.

When the area that maximizes $R_P^{(S)}$ (step 2) is located at the edge of the system, all particles getting out of and into the corresponding area (step 3) are supposed to move to and from, respectively, its one neighbor.

Figure 4 summarizes the simulation results. We assumed two kinds of profiles for TARGET, as shown in Figures 4A (TARGET A) and B (TARGET B). These are given below:

$$\text{TARGET A: } \sin(2\pi x/N \times 8) \text{ and TARGET B: } 0.8 \times [\sin(2\pi x/N \times 4) + \sin(2\pi x/N \times 8)]. \quad (15)$$

We assumed that the total number of pixels at the finest scale ($S = 0$) in the horizontal direction is given by $N = 2^{10}$ (= 1024).

The degree of adaptation of GRIPPER to TARGET is evaluated by

$$R_a = \sum \left| h^{(0)}(x) - \overline{h^{(0)}(x)} \right| / N. \quad (16)$$

This means that the average of the absolute values of the deviation from the average difference between GRIPPER and TARGET, which is known as $R_a$, is used as a measure to quantify the surface roughness in the literature (Gadelmawla et al., 2002; Naruse et al., 2013). The initial values of $R_a$ with respect to the profiles shown in Figures 4A and B are the same.

Figure 4C shows the evolution of the profile of GRIPPER when the object is given by TARGET B. As the time elapses, the shape of GRIPPER becomes closer to TARGET B. The metric $R_C^{(S)}$ increases at coarser scales compared with finer scales in the early stages of the adaptation because GRIPPER does not include fine-scale structures. In contrast, as time elapses, shape changes at finer scales are induced. Numerically, the demonstration shown in Figure 4C considered six different scales: $S = 2, 3, …, 7$. That is, $S_{min}$ and $S_{max}$ defined in step 1 were 2 and 7, respectively, and the sizes of local areas at each scale ($2^S$) were given by 4, 8, 16, 32, 64, and 128. We repeated the evolution of GRIPPER 1000 times, beginning with the same initial condition (flat surface), and examined the average values in the following analysis.

As shown by the red curve in Figure 4D, $R_a$ decreased as time elapsed. We configured $S_{min}$ to increase, which means that the minimum physical scale considered in the adaptation dynamics was increased, in order to examine the extent of adaptation. This physically corresponds to an increase in the size of the elemental particles contained in the glove of the gripper. The green, blue, and cyan curves in Figure 4D represent the evolution of $R_a$ when $S_{min}$ was 3, 4, and 5, respectively. The achievable minimum $R_a$ increased with the minimum scale, which means that the fit between GRIPPER and TARGET was not perfect. Note that increasing the minimum scale obtained certain $R_a$ faster than at larger scales. For example, until the time cycle of 375, $R_a$ decreased most rapidly with $S_{min} = 5$. Likewise, until the cycles of 888 and 1398, $R_a$ decreased fastest with $S_{min}$ of 4 and 3, respectively. It is also noted that $R_a$ decreases in the same manner until around cycle 200, regardless of the given $S_{min}$: this is because the adaptation in the initial phase is dominated by the coarsest scale. A related aspect is further elaborated shortly below.

These results are accounted for by the mathematical framework presented by the category of mobility: the richness of the category of mobility with regard to the minimum physical scale of the model ($S_{min}$). A small $S_{min}$ (e.g., 3) obtains a very small $R_a$; however, this can mean that the robot is too soft, either in that the possible friction between the robot and target is weak or that it takes more time to reach a steady state (constant $R_a$ value). Thus, the concept of an effectively soft robot is demonstrated.

Indeed, the number of states and transitions between states in the numerical model increases exponentially as the physical scale of interests becomes smaller, as shown by the red circles and the blue squares in Figure 4E, respectively. In this study, for simplicity, the number of states at scale $S$ was estimated as the number of arrangements of a total of $N / 2^S$ blocks over a horizontally expanding area of $N / 2^S$. A transition between states is defined such that a block, among $N / 2^S$ blocks, moves toward either the right- or left-hand side. As shown in Figure 4E, with $S = 3$ or at a finer scale, the number of states and transitions reaches huge numbers, in the order of $10^{75}$ and $10^{77}$, respectively; whereas at a coarser scale with $S = 7$, they give smaller figures, in the order of $10^3$ and $10^4$, respectively. With these calculations, it is clear that there is an overabundance of states and transitions, which was argued through natural transformation and the definition of "effectively soft robot" in the previous section.

Furthermore, the dynamics of the particle flow behaves differently depending on the shape of the object. Figure 4F compares the evolution of $R_a$ regarding TARGET A and B, which are marked by red and blue curves, respectively. $R_a$ decreased faster with TARGET A than B. Although the initial value of $R_a$ was the same for A and B, they had different spatial frequencies, as shown in equation (15). To examine the underlying behavior, Figure 4G shows the number of particles transferred between the regions. For example, when the metric $R_P^{(S)}$ was maximized at the scale $S = 5$, the number of particles moving in the system was $2^5$ (= 32). Specifically, here, the number of transferred particles at cycle $t$ was evaluated by the average of the number of transferred particles from cycle $t-9$ to $t$ for the purpose of observing the overall evolution. Both TARGETS A (red) and B (blue) behaved similarly in the initial phase until around cycle 100. During this phase, the number of transferred particles was 64, which means that $R_P^{(S)}$ was maximized at the scale $S = 6$; this coincides with the identical initial trajectories of *Ra* observed in Figure 4D.

When the cycle was around 200, the number of moving particles decreased significantly. This corresponds to the situation where the adaptation between the target and gripper progressed only at the finer scales. This is physically natural because TARGET A had a simple periodic structure; hence, the selected physical scale was basically monotonically decreasing. In contrast, TARGET B exhibited quite different sequences. Because TARGET B had two spatial frequencies, the physical scale that maximized $R_P^{(S)}$ differed depending on the shape in each cycle. Further, especially with TARGET B, the number of particles transferred between the sites exhibited oscillatory behavior from around cycle 100. That is, the physical scale that gives the maximum fitness figure between the gripper and the target may occasionally increase even though *Ra* is monotonically decreasing. These results show the richness of the category of mobility for soft robots where the scale of control is adaptively and autonomously configured. This is another aspect of the concept of an effectively soft robot as described by the theory.

Before finishing the discussions, we add a few remarks. First, while the present study specifically deals with universal gripper as a typical example, the categorical approach in the present paper has general versatile applicability to various sorts of soft robots in such forms as soft-legged robot, octopus' arm, among others. We emphasize that our approach is surely applicable to other various platforms on the basis of the following reason.

The point is that the effectively softness of a robot can be defined whenever we can define the categories of mobility of the composite systems of the robot and the targets. The composite systems in the case of universal gripper is the composite systems of the gripper and the targets to be grasped. The robot is called effectively soft if its control on any target of certain kind is effectively soft. In the case of the walking soft-legged robot, for example, we can define the effectively softness by considering the composite systems of the robot and the system of certain kind on which the robot will walk.

Conversely, let us consider a continuously deformable soft material under gravity. Here the notion of effectively softness is *not* well-defined, since it essentially depends on the choice of the class of systems to be targets. Once the class of target systems is specified, in other words, the *purpose* of the robot such as grasp or walk is chosen, then we can define effectively softness in our category theoretic framework.

Finally, we discuss the contribution of the present study and future works. One may ask the following question: Is this study a design theory or an architecture for soft robots? We here state that it is the former. We consider that there are huge descriptive benefits in the design of soft robots where the essence of category theory, especially natural transformation, is greatly utilized. Meanwhile, the benefits in synthesis or optimization in the design process of soft robots are unfortunately *not directly* or *not explicitly* derived by the categorical theory itself, at least at this stage of research. Certain specific numerical models, such as the one demonstrated in our paper, taking the universal gripper example, should be accommodated. However, we would like to emphasize that, based upon the concept of categorical equivalence, certain specific models can be constructed in a straightforward manner because natural transformation suggests the kinds of aspects that should be highlighted and addressed in the analysis or design processes. Indeed, the derivation of the scale-dependent fitness measure between the gripper and the object demonstrated in the numerical model, as well as the calculation of the estimated number of states and transitions, may *not* be an evident nor obvious examination without categorical understanding of the universal gripper. It is specifically in this sense that the impact on synthesis or optimization aspects of design is partially argued in this paper.

From these considerations, an important future study is construction of theories that provide visible and engineering insights and give fundamental limits for soft robotics from category theory. Indeed, Naruse *et al.* proposed "short-exact-sequence-based time" based on homological algebra and triangulated category to account for the maximum operating speed in a photon-based solution searching system (Naruse et al., 2017), which addresses the fact that an operation that is too fast does not lead to proper results. A unification of the notions of category of mobility and the theorems known in the triangulated category would be an interesting future study. Of course, we should examine the applicability of the categorical approach by practicing on a variety of concrete engineering systems, not just the universal gripper; for example, reservoir computing, which has been recently implemented in a variety of media such as soft robots (Nakajima et al., 2013) and chaotic lasers (Nakayama et al, 2016).

## 4. CONCLUSION

We proposed a mathematical foundation for soft robotics based on category theory. The category of mobility with the notions of functors and natural transformation provides a rigorous formulation for soft robots and their interactions with the target object or environment. The difference with hard robots and the effectiveness of a soft robot were mathematically described: the former and the latter are based on isomorphism and categorical equivalence, respectively. Natural transformation provides rich transitions between states providing deformable, autonomous, adaptive behavior of soft robots, whereas a one-to-one correspondence between the state and transitivity of states restricts the rigid movements of hard robots. In the meantime, overabundant transitivity could lead to unstable behavior; for example, a surface profile that is too smooth may not realize successful grasping of target objects.

We introduced the notion of effectively soft robot in order to realize intended functionality with the notion of critical state. As an application of the theory, a model system and analysis were presented to examine the adaptation behavior observed in universal grippers. The scale dependency of the elemental particles contained in the gripper was observed to agree with the theoretical prediction; indeed, including an overly small scale yields slower adaptation. It should be noted that the scale-dependent adaptation measure was inspired by the notion of effectively soft robot in the theory. The number of states and transitions therein was numerically estimated, where smaller scale yield huge numbers. The autonomous adaptation behavior was demonstrated where different state transitions were clearly observed depending on the profiles of the object where even oscillatory behavior was observed.

The power of category theory in its descriptive aspect is clear. The benefits in synthesis and optimization have been partially demonstrated in the present study through the modeling and analysis of the universal gripper. As discussed at the end of Section 3, there is a wide range of exciting future studies ranging from constricting general theory, which we can provide benefits to in concrete synthesis processes, not just the exact mathematical description to exercising various application systems, including reservoir computing.

**AUTHOR CONTRIBUTIONS**

**ACKNOWLEDGMENTS**

We acknowledge K. Nakajima for variable discussions regarding theories and implementations of soft robots.

**FUNDING**

This work was supported in part by the Core-to-Core Program A. Advanced Research Networks, Grants-in-Aid for Scientific Research (A) (JP17H01277) from Japan Society for the Promotion of Science, CREST program (JPMJCR17N2) from Japan Science and Technology Agency.

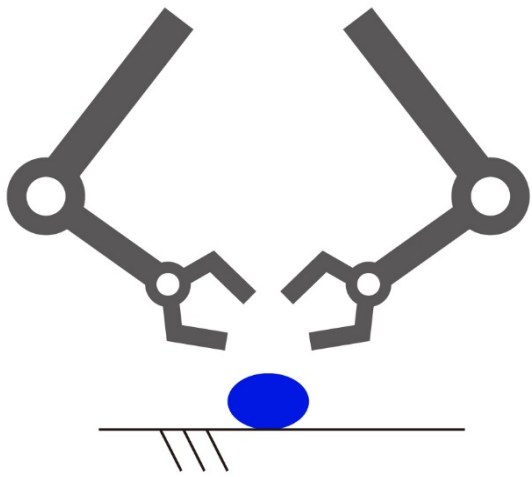 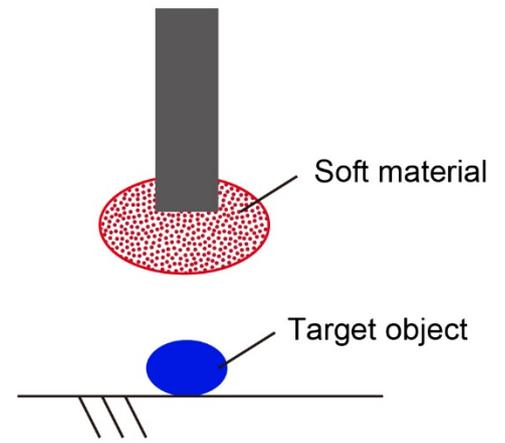

**FIGURE 1 | Hard and soft robots. (A)** Hard robots require precision control of rigid joints to grasp an object, whereas **(B)** a soft robot autonomously adapts its shape to the object.

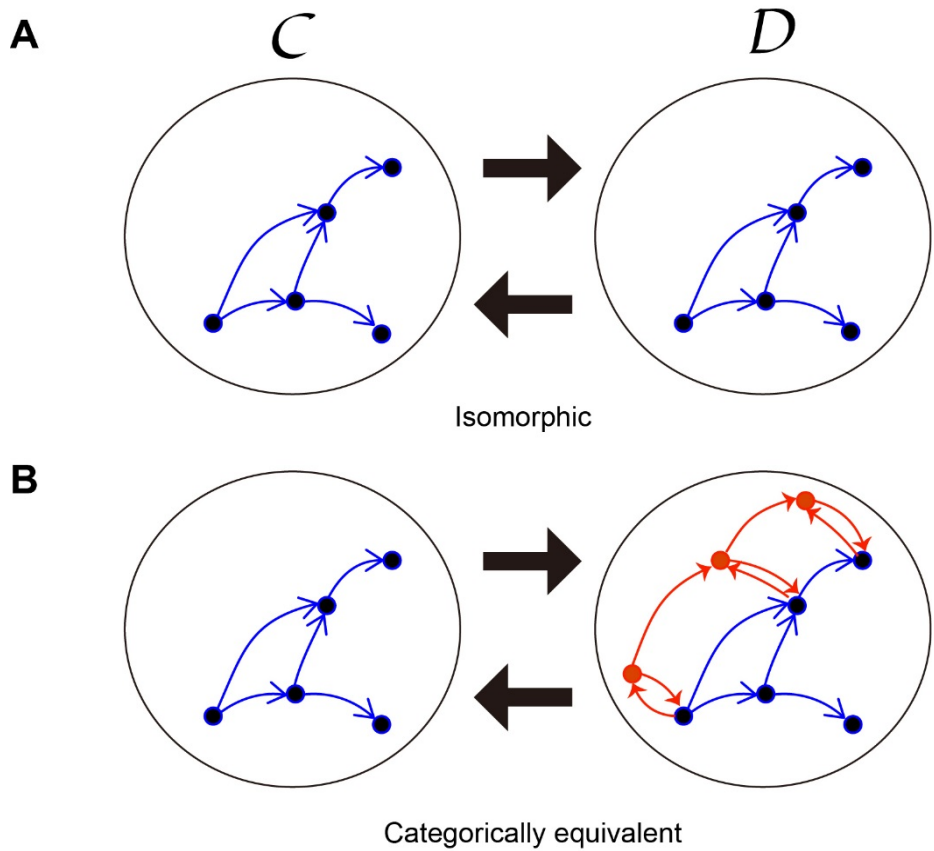

**FIGURE 2 | Mathematical understanding of the difference between hard and soft robotics. (A)** *Isomorphisms* describe the underlying principle to describe hard robots: a coherent one-to-one correspondence between the state and transitivity of states of the composite system and robots during interaction. **(B)** On the other hand, *categorical equivalences* represent the architecture of soft robots: abundant degree-of-freedoms or versatile possibilities are accommodated during interaction.

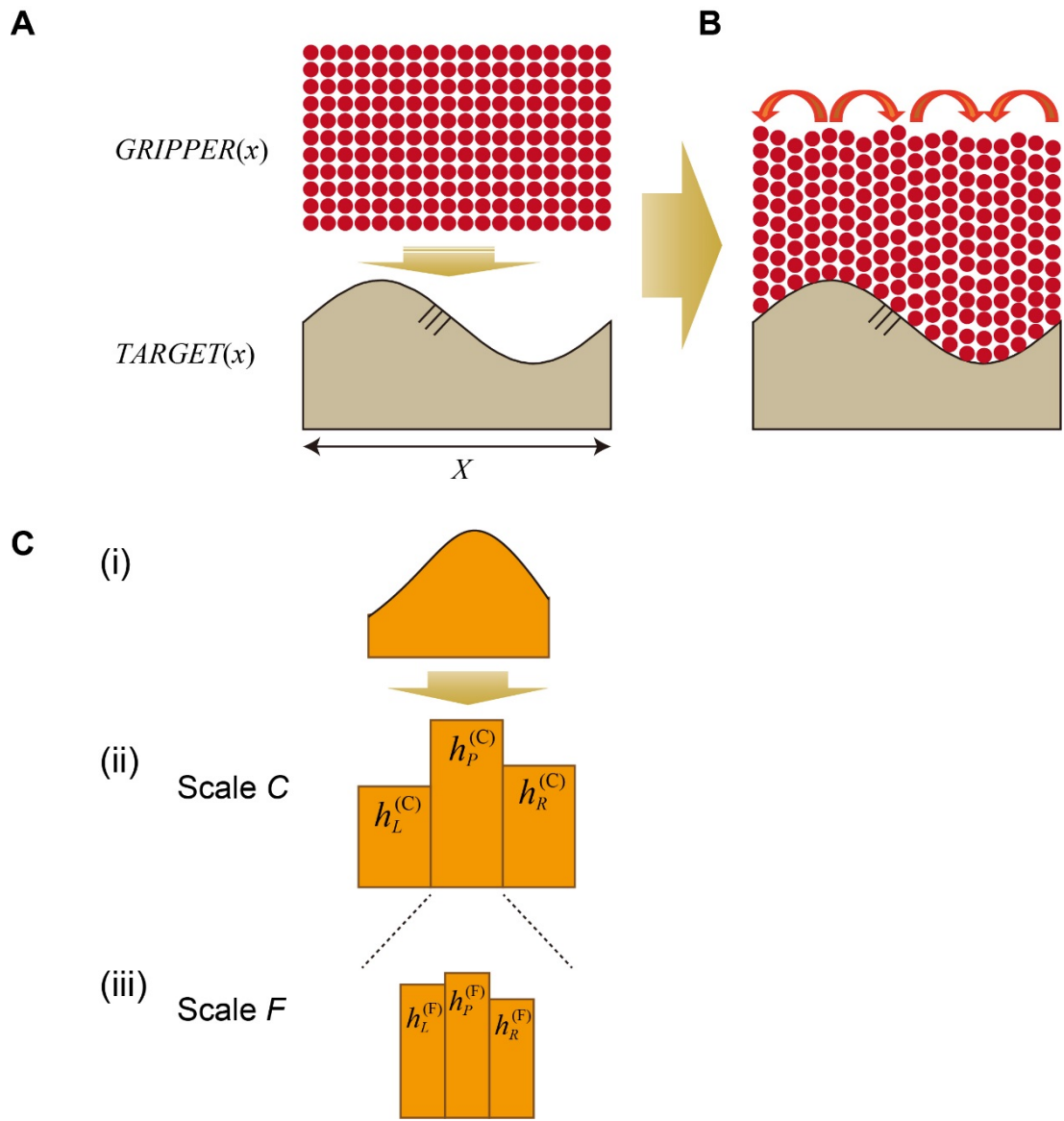

**FIGURE 3 | Model of the autonomous adaptation with soft robotics. (A)** Schematic diagram of a soft-material-based hand *GRIPPER*(*x*) and the target object with an arbitrary shape *TARGET*(*x*). **(B)** The internal microstructure is autonomously reconfigured for adaptation. **(C)** Scale-dependent characterization: (i) original structure, (ii) coarse-scale structure, and (iii) fine-scale structure.

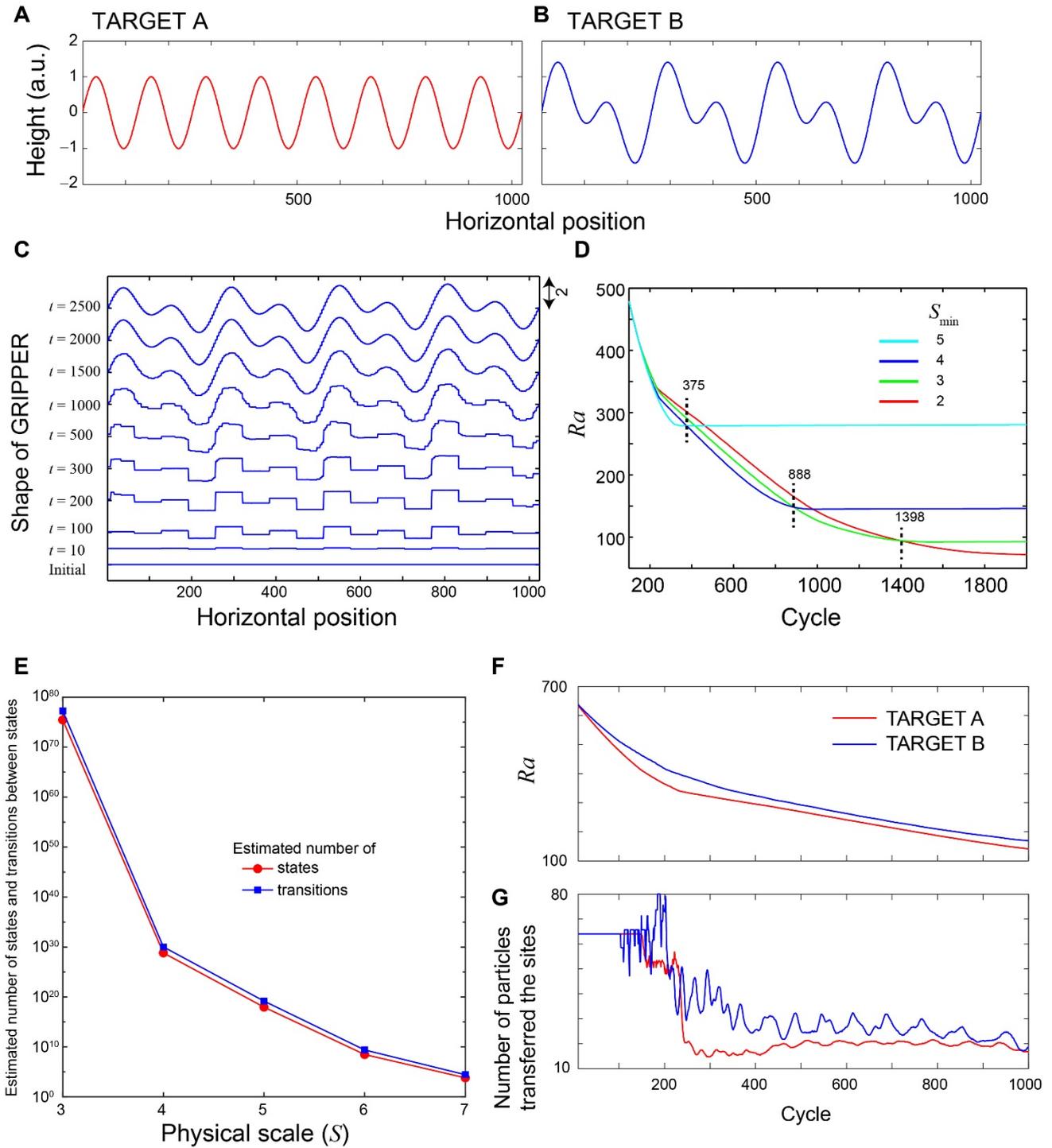

**Figure 4 | Demonstration of the scale-dependent adaptation with soft robotics. (A B)** Assumed profiles of the target object. Both **(A)** and **(B)** have the same surface roughness (*Ra*), but **(A)** contains a single spatial frequency while **(B)** consists of two frequencies. **(C)** From the initial flat surface, the shape of the gripper adaptively changes to fit to the object. **(D)** Evolution of the fitness figure *Ra* as a function of time. Depending on the minimum scale to be considered ($S_{min}$), the dynamics exhibit

different characteristics. **(E)** Estimated number of states and transitions between states at different scales of the model. **(F, G)** Comparison of the shape change of the soft robot during the adaptation to different target objects [**(A)** and **(B)**]. The evolution of *Ra* **(F)** and the movement of internal materials **(G)** differ significantly depending on the target object, although the surface roughness is the same.